\documentclass[journal]{IEEEtran}
\usepackage{times}
\usepackage{epsfig}
\usepackage{graphicx}
\usepackage{amsmath}
\usepackage{amssymb}
\usepackage{subcaption}

\usepackage{array}
\usepackage{mathrsfs}
\usepackage{booktabs}
\usepackage[pagebackref=true,breaklinks=true,letterpaper=true,colorlinks,bookmarks=false]{hyperref}
\usepackage{amsmath}
\usepackage{amssymb}
\usepackage{algorithmic}
\usepackage{color}

\ifCLASSINFOpdf
\else
\fi
\begin{document}

%
\title{CP-Net: Contour-Perturbed Reconstruction Network for Self-Supervised Point Cloud Learning}
%
%
%

\author{Mingye Xu,
	    Yali Wang\textsuperscript{ \footnotemark* }
        Zhipeng Zhou,
        Hongbin Xu,,
        and Yu Qiao\textsuperscript{ \footnotemark*} ,~\IEEEmembership{Senior Member,~IEEE}
\thanks{
M. Xu is with  University of Chinese Academy of Sciences.
}
\thanks{
M. Xu, Y. Wang and Y. Qiao are with Shenzhen Institutes of Advanced Technology, Chinese Academy of Sciences, China.
}
\thanks{
Z. Zhou is with the Algorithm Expert in Alibaba Damo Academy.
}
\thanks{
H. Xu is with South China University of Technology.
}
\thanks{ * The corresponding authors. }


}

%
%

\markboth{}%
{Shell \MakeLowercase{\textit{et al.}}: SUBMITTED TO IEEE TRANSACTIONS ON MULTIMEDIA}

%



\maketitle
\begin{abstract}

Self-supervised learning has not been fully explored for point cloud analysis.
Current frameworks are mainly based on point cloud reconstruction. 
Given only 3D coordinates,
such approaches tend to learn local geometric structures and contours,
while failing in understanding high level semantic content.
Consequently, they achieve unsatisfactory performance in downstream tasks such as classification, segmentation, etc.
To fill this gap, we propose a generic Contour-Perturbed Reconstruction Network (CP-Net),
which can effectively guide self-supervised reconstruction to learn semantic content in the point cloud,
and thus promote discriminative power of point cloud representation. 
First,
we introduce a concise contour-perturbed augmentation module for point cloud reconstruction. 
With guidance of geometry disentangling, 
we divide point cloud into contour and content components.
Subsequently, 
we perturb the contour components and preserve the content components on the point cloud.
As a result,
self supervisor can effectively focus on semantic content,
by reconstructing the original point cloud from such perturbed one.
Second, 
we use this perturbed reconstruction as an assistant branch,
to guide the learning of basic reconstruction branch via a distinct dual-branch consistency loss.
In this case,
our CP-Net not only captures structural contour but also learn semantic content for discriminative downstream tasks.
Finally, 
we perform extensive experiments on a number of point cloud benchmarks. 
Part segmentation results demonstrate that our CP-Net (81.5\% of mIoU) outperforms the previous self-supervised models, and narrows the gap with the fully-supervised methods.
For classification, we get a competitive result  with the fully-supervised methods on ModelNet40 (92.5\% accuracy) and   ScanObjectNN  (87.9\% accuracy).
The codes and models will be released afterwards.

\end{abstract}

\begin{IEEEkeywords}
3D point cloud analysis, Unsupervised learning,  Classification, Part segmentation
\end{IEEEkeywords}

%
\IEEEpeerreviewmaketitle

\section{Introduction
\label{Sec_Intro}}


\IEEEPARstart{P}oint cloud analysis has gradually become an important problem for understanding 3D world,
resulting from its wide applications in robotics, AR/VR, autonomous driving, etc  \cite{rusu2008towards,qi2018frustum,9154552,7904634,9234727}.
Current mainstream m                                                                                              ethods  of point cloud analysis are mainly driven by fully supervised deep learning \cite{Charles_2017,qi2017pointnet++,Xu_2020_GSNet,xu2018spidercnn,wang2018dgcnn,Liu2019RSCNN,thomas2019kpconv}.
However, these  methods require a large number of manual annotations, which could be expensive and infeasible for practical applications.
Therefore, it is desirable to obtain discriminative representations of 3D point cloud in a self-supervised manner. 

\begin{figure*}[t]
\begin{center}
   \includegraphics[width=1\linewidth]{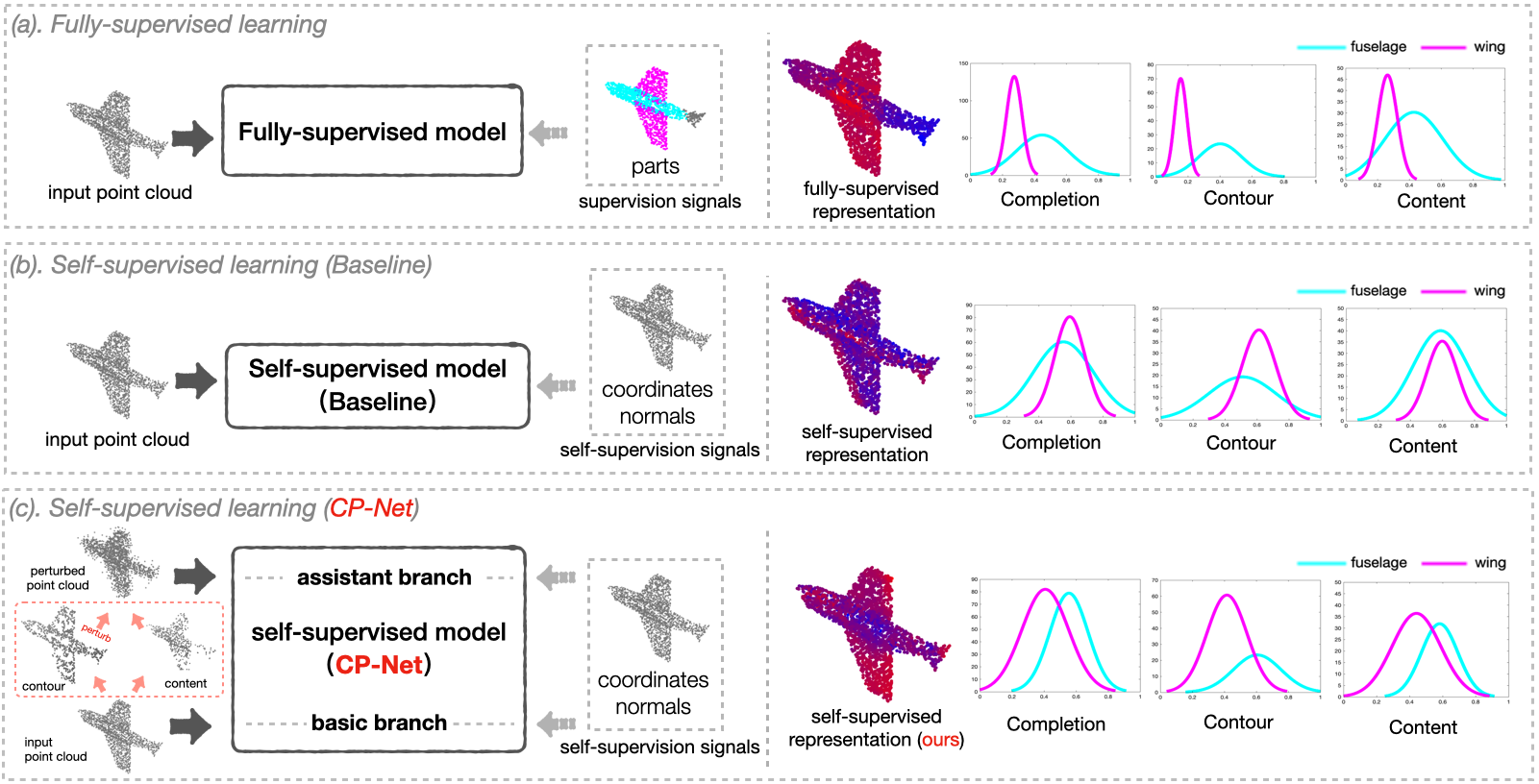}
\end{center}
   \caption{Point cloud  representation learning of fully supervised method (a)  and  self-supervised  baseline method (b)  and our CP-Net (c).
   	On the left is the basic structure of these models.
   The right parts are the feature representation and distribution curves of the semantic parts (fuselage and wing). 
   (a): It can be observed that fully-supervised feature representation can be well distinguished  among different parts. 
   (b): Self-supervised methods can learn contours well, but not content. For example, the distribution curves of the wing and the fuselage overlap in the content.
   (c): The self-supervised feature distribution of our CP-Net shows the distinction both in contour and content components.
 }
\label{fig_introduction_1}
\end{figure*}

The current self-supervised methods are mainly based on pretext tasks provided by generation or reconstruction \cite{gadelha2018multiresolution,han2019multi,li2018pointgan,liu2019l2g,zhao2019_3DcapsNet,rao2020pointGLR,shoef2019pointwise,yang2018foldingnet,zhang2020UFF,hassani2019unsupervised}.
However, their performance is far from that in the fully-supervised methods.
To fill this gap, we investigate what is the underlying problem in the self-supervised point cloud learning.
As Figure \ref{fig_introduction_1} shows, we take the point cloud segmentation task of an airplane for illustration.
We first visualize point cloud by the corresponding feature response in different methods.
It can be observed that,
the self-supervised reconstruction is good at learning structural contours, 
but fails in distinguishing semantic content,
e.g.,
the contours of $fuselage$ and $wing$ are clear,
while 
their contents are confused with same feature responses.
This further motivates us to investigate self-supervised point cloud representations,
in terms of contour and content.
Specifically,
we use geometry disentangle module \cite{xu2020GDA} to divide point cloud into contour and content components,
and show the feature response distribution of $fuselage$ and $wing$ on 
the complete point cloud, 
contour and content components in Figure  \ref{fig_introduction_1}.
As expected,
the feature distribution of different semantic parts can be easily separated in the contour components,
while 
they are heavily overlapped in the content components.
It clearly indicates that,
self-supervised reconstruction lacks capacity of distinguishing semantic parts in the content components.

To address this difficulty,
we propose a generic Contour-Perturbed reconstruction network (CP-Net),
which can effectively guide self-supervised reconstruction to pay more attention to discriminative object content,
based on point cloud disentangling.
First,
we geometrically divide a point cloud into contour and content components,
and then augment it by perturbing contour and preserving content.
By reconstructing original point cloud from the contour-perturbed one,
we can effectively force self-supervisor to learn semantic content.
Second,
we build up a weight-sharing dual branch structure to boost point cloud representation learning.
Since the basic branch is good at learning structural contour while ignoring semantic content,
we leverage such perturbed reconstruction as an assistant branch of the original reconstruction task.
Via designing a novel dual-branch consistency loss,
we can progressively use the assistant branch to guide the basic branch for learning semantic content.
In this case,
the basic branch can capture easy-to-learn contour as well as exploit necessary-to-learn content,
which enhances point cloud representation of self-supervised reconstruction for discriminative downstream tasks.
Finally, the experiments and visualization analysis in Sec. \ref{Sec_analysis}, \ref{Sec_exp} demonstrate the effectiveness, robustness and generalization ability of our method.
The main contributions are summarized as follows,
\begin{itemize}
	\item We propose a generic Contour-Perturbed Reconstruction Network (CP-Net) for self-supervised point cloud feature learning, which can effectively learn the discriminative representation both on structural contour and semantic content components.
	\item We explore an effective contour-perturbed augmentation  module to perturb  contour components of point clouds, which is used to force the assistant  branch to learn the semantic content information.
	\item 
	We introduce a multi-scale dual-branch consistency loss, which can bring the the corresponding features of the basic and assistant branches closer.
	Then the   basic branch can pay more attention to the semantic content information by the guidance of  assistant  branch.
	\item Experiments demonstrate that our method  outperforms the self-supervised methods and narrows the gap between unsupervised and supervised models in part segmentation task (81.5\% mIoU of ShapeNetPart).
	For classification, our self-supervised method gets a comparable result with the fully-supervised methods on ModelNet40 (92.5\% accuracy) and ScanObjectNN (87.9\% accuracy).
\end{itemize}

\begin{figure*}[t]
\begin{center}
   \includegraphics[width=1\linewidth]{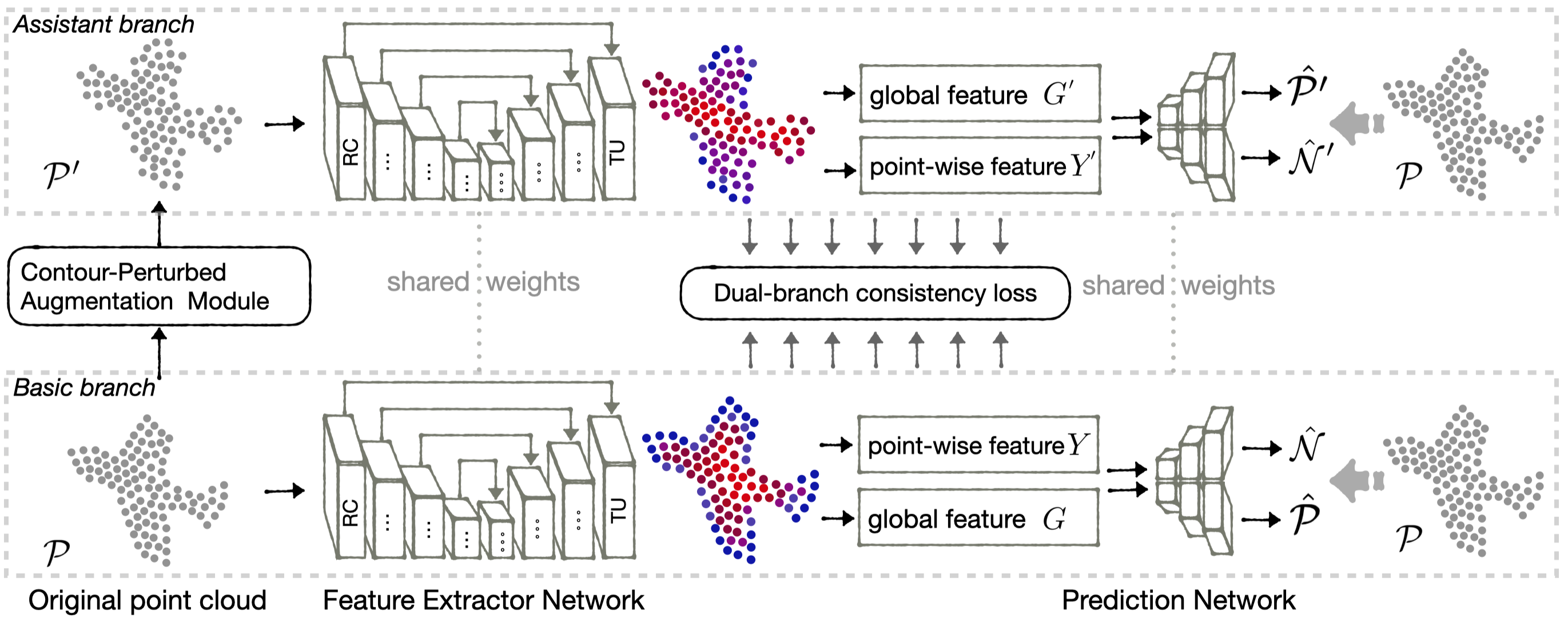}
\end{center}
   \caption{Network architecture of our CP-Net. "RC" means RS-Conv module, "TU" means transition up module. The basic branch uses the original point cloud to reconstruct its coordinates and normal vectors.  While the assistant branch takes the contour-perturbed point cloud as input and reconstruct the original point cloud.  The feature extractor network and prediction network from two branches share the parameters. 
     }
\label{fig_method_1}
\end{figure*}

\section{Related Work}

\subsection{Supervised Learning on 3D Point Clouds}
Recently 3D point cloud analysis has enjoyed some remarkable progress for various downstream tasks,
PointNet \cite{Charles_2017} and DeepSet \cite{Zaheer2017Deep} are pioneering architectures that directly process point cloud. 
Their basic idea is to learn a spatial encoding of each point and fusing all individual point features to a global signature with max pooling.
Though efficient, the local geometry structures are not sufficiently captured. 
To remedy this, PointNet++ \cite{qi2017pointnet++} extracts local features capturing fine geometric structures from neighbors through a hierarchical grouping architecture.
Some subsequent works such as PointCNN \cite{li2018pointcnn}, PointConv \cite{wu2019pointconv} and RSCNN \cite{Liu2019RSCNN} also focus on the extraction of local geometric features. 
To capture the holistic geometric information more efficiently, GS-Net \cite{Xu_2020_GSNet} groups distant points with similar and relevant geometric information and aggregates features from neighbors in both Euclidean space and Eigenvalue space.
DGCNN \cite{wang2018dgcnn} reconstructs the k-nn graph using nearest neighbors in the features space  dynamically.
Although these supervised methods push state-of-the-art of point cloud deep learning with the help of extensive supervised signals, the generalization ability may be limited by the supervised learning mechanism.
Therefore, it is desirable to obtain features in an unsupervised manner and obtain the general representation of 3D point clouds.

\subsection{Unsupervised Learning on 3D Point Clouds}
In order to produce a semantic latent space without relying on  annotations, the unsupervised network is trained to perform the tasks based on some information obtained from the point cloud itself.
Based on this, recent  self-supervised approaches design various pre-tasks such that models need to learn useful information from data itself \cite{bachman2019learning,doersch2015unsupervised,doersch2017multi,henaff2020data,tian2019contrastive}.
Several prior works have attempted on learning representation of point cloud without human supervision \cite{rao2020pointGLR,yang2018foldingnet,zhao2019_3DcapsNet,zhang2020UFF,xie2020pointcontrast,hassani2019unsupervised}.
FoldingNet \cite{yang2018foldingnet} trains an end-to-end  auto-encoder that consumes unordered point clouds directly by reconstruction from the point cloud itself. 
PointGLR \cite{rao2020pointGLR} focuses on reasoning between local and global representations.
GraphTER \cite{gao2020graphter} proposes graph transformation equivalent representation learning to extract unsupervised representations.
Chen et. al. \cite{chen2021shape}   destroies certain local shape parts of an object, and then segment points that belong to distorted parts via a point cloud network. 
Different from all these previous works,
 we explore a dual-branched self-supervised learning framework (CP-Net), which can guide the self-supervisor to learn the  discriminative representation  both on  semantic content and structural contour. 
Moreover, the self-supervised representation of our CP-Net is more friendly to the downstream tasks.
 

\section{Method}
This section will introduce our proposed CP-Net in detail.
First, we elaborate on the overall framework.
Then, we introduce our contour-perturbed augmentation    module for assistant branch. 
Finally, we describe the loss terms in our CP-Net.

\subsection{Overall Architecture of Our CP-Net}


Our CP-Net is a generic dual-branched network, which consists of assistant branch and basic  branch.
The assistant branch is used to learn discriminative representation on semantic content, 
while the basic branch   preserves the discriminative representation of the  structural contour.
Since both branches are used for point cloud reconstruction, they share the  weights of feature extraction network and prediction network.
Feature extraction network  is designed to obtain the global feature and point-wise feature.
Prediction network reconstructs the coordinates and estimate the normal vectors.
By introducing  dual-branch consistency loss as feature consistency regularization, we can leverage the assistant branch to guide the basic branch for distinguishing content information of point clouds.

As shown in Figure \ref{fig_method_1}, we consider a 3D point cloud with $N$ points as the input.
Generally,  point cloud contains 3D coordinates 
$\mathcal{P}=\{{p}_{1}, \ldots, {p}_{N}\}$,
and normal vectors 
$\mathcal{N}=\left\{{n}_{1}, \ldots, {n}_{N}\right\}$.
The input of basic branch is the original point cloud coordinates $\mathcal{P}$, while the input of assistant branch is the perturbed point cloud coordinates $\mathcal{P}'$ by contour-perturbed augmentation  module. 
This extractor network recieves point cloud coordinates as input, and  outputs the point-wise features $Y$ and  global features $G$.
The reconstructed coordinates and normal vectors of prediction network can be defined as $\hat{\mathcal{P}}$ and $\hat{\mathcal{N}}$.



\textbf{Feature Extraction Network.}
Like PointNet++ \cite{qi2017pointnet++}, we use a hierarchical structure to learn point cloud feature progressively with skip connections.
Specially, at $l$-th level of encoder, point set is downsampled by using iterative furthest point sampling (FPS) to produce a new point set $\mathcal{P}^{l} \subset{\mathcal{P}^{l-1}}$ with $N^{l}$ points from $N^{l-1}$ points.
Meanwhile we  extract the  point-wise feature $f^{l}_{i} \subset F^l$ by applying the RS-Conv  \cite{Liu2019RSCNN}  for each point ${p}_i^l \subset \mathcal{P}^l$.
For corresponding $l$-th level of decoder, the input feature is ${Q}^{l}$,  we  use the transition-up module \cite{zhao2020pointtransformer} to propagate the points feature:
\begin{equation}
Q^{l-1}={MLP}\left({MLP}({\xi( Q^{l}})) + MLP( F^{l-1})  \right)
\end{equation}
where $\xi$ is the point interpolation function \cite{qi2017pointnet++}.
Moreover, we  also use the transition-up module to propagate the  feature   ${Q}^{l}$ of each level $l$ to ${Y}^{l}$ from $N^l$ points to the original $N$ points:
\begin{equation}
    {Y}^{l}={MLP\left( {{\xi}( Q^{l}} )\right)}
\end{equation}
Then we concatenate them together as the  point-wise feature 
$Y = Concate[{Y}^{1},..., {Y}^{L}]$
, where $L$ is the layers of feature extractor network.
While the global feature $G$ is obtained by a symmetric aggregation function (e.g., max pooling, ...) operating on the point-wise feature.

\begin{figure}[t!]
	\begin{center}
		\includegraphics[width=1\linewidth]{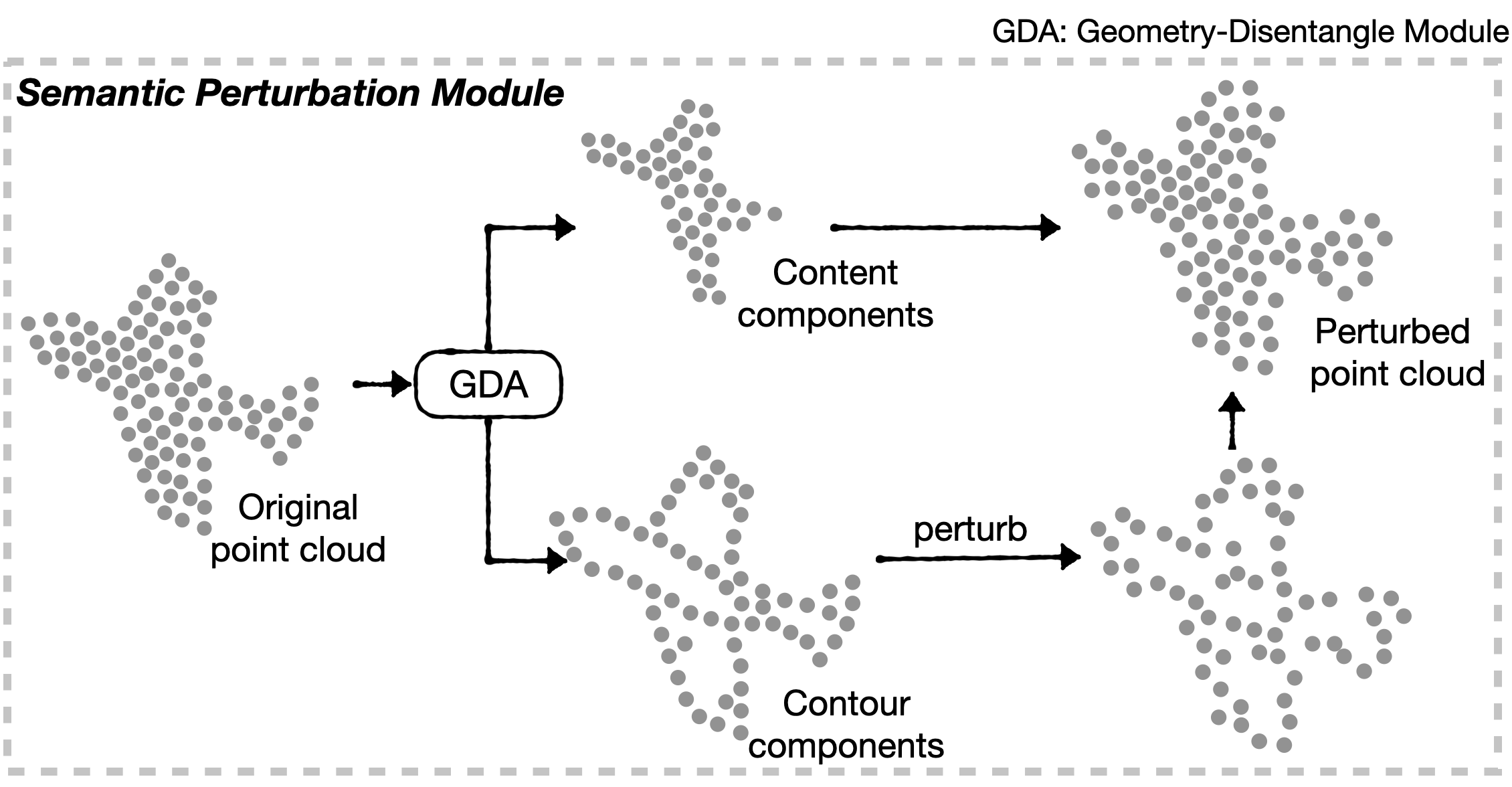}
	\end{center}
	\caption{The process of contour-perturbed augmentation  module.}
	\label{fig_spm}
\end{figure}

\textbf{Prediction Networks.}
The self-supervised prediction networks consist of  normal prediction network and point cloud reconstruction  network. 
The normal prediction network is used to enhance point cloud representation with geometry structure information.
We can take the concatenation of the global feature $G$,  original point coordinates ${p}_i$  and  point-wise feature $y_i$ from $Y$ as input, then obtain the estimated normal $\hat{{n}}_{i} \subset  \widehat{\mathcal{N}}$ through a shared light-weight MLP and $l^2$ normalization operation $\amalg$:
\begin{equation}
\hat{{n}}_{i} = \amalg({MLP}\left({p}_{i} \oplus {y}_{i} \oplus G\right)) 
\end{equation}
where $\oplus$ is the concatenation operator, and $\hat{{n}}_{i} \in \mathbb{R}^{3} $.

The self-reconstruction network is used to recover the coordinate information of the original point cloud.
Referring to FoldingNet \cite{yang2018foldingnet}, by incorporating a standard two-dimensional grid, 
we can  deform the reconstructed  coordinates with the guidance of the  global feature $G$.
The self-reconstruction network contains  two consecutive 3-layer MLPs. 
Specially,
before we feed the global feature $G$ into this network,  we replicate it $N$ times as $\hat{G}$ ,
then concatenate it with an  matrix $I \in \mathrm{R}^{N \times 2}$ which contains the $N$ grid points on a square centered at the origin \cite{yang2018foldingnet}.  
The reconstructed point cloud $\hat{\mathcal{P}}' \in  \mathrm{R}^{N \times 3}$ can be obtained by the following operation:
\begin{equation}
	\hat{\mathcal{P}}' =  MLP(\hat{G}\oplus  MLP (\hat{G}\oplus  I))
\end{equation}
where $\oplus$ is the concatenation operator.

\subsection{Contour-Perturbed Augmentation  Module }

As mentioned in the Sec.\ref{Sec_Intro},
self-supervised reconstruction mainly focuses   on structural contours, while ignoring the discriminative  content information of point cloud.
To tackle this problem, we design the contour-perturbed augmentation  module  for the assistant branch in Figure \ref{fig_spm}. 
We have observed that the geometry-disentangle module \cite{xu2020GDA} can decompose the point cloud into the structural contour information and semantic content information.  
Inspired by this, we can design our contour-perturbed augmentation  module  to perturb the point cloud.


First,
for the input point cloud $\mathcal{P}$ with $N$ points, we  construct the point graph with the eigenvalues which represent the graph frequencies. 
Second, we collect the contour components $\mathcal{P}_{s} \in \mathbb{R}^{M \times 3}$ and content components $\mathcal{P}_{g} \in \mathbb{R}^{M \times 3}$ through the graph filters \cite{xu2020GDA} on the constructed graph, where $M = N/2$. 
The points in the contour components $\mathcal{P}_{s}$ are easier to describe the local geometric structural information of the point cloud,
while the content points $\mathcal{P}_{g}$ can highlight the relatively common semantic information.
Third, we perturb the contour points $\mathcal{P}_{s}$ with normal distributed noise $ \Delta \in \mathbb{R}^{M \times 3}$,
then we concatenate the perturbed contour points $\mathcal{P}_{s}' = \mathcal{P}_{s} + \Delta \in \mathbb{R}^{M \times 3}$ with original content points $\mathcal{P}_{g}$ as the perturbed point cloud $\mathcal{P}' = \mathcal{P}_{s}' \oplus \mathcal{P}_{g}$, where $\mathcal{P}'\in \mathbb{R}^{N \times 3}$.
Finally, we use the perturbed point cloud  $\mathcal{P}'$ to reconstruct the original point cloud in the assistant branch,
which makes this branch  pay more attention to the semantic content information.

Besides of these methodology insights,
our experiments in Table \ref{Anysis_contrast_pointcloud} can also prove our statement,
it clearly shows that content components are harder to learn in the self-supervised manner.
Moreover, we also consider other point cloud decomposition schemes, such as point cloud clustering, spatial domain decomposition, etc.  Specific results and analysis are presented in Table \ref{Anysis_contrast_pointcloud}.


\subsection{Loss Terms}
To learn our model effectively,
we introduce three training losses:
\begin{equation}
    Loss = \mathcal{L}_{guid} + \mathcal{L}_{recon} + \mathcal{L}_{normal}
\end{equation}
where the dual-branch consistency loss $ \mathcal{L}_{guid}$ is used to guide the basic branch to learn the semantic content representation with assistant branch.
$\mathcal{L}_{recon}$ and $\mathcal{L}_{normal}$ are  widely used  self reconstruction losses.

\textbf{Dual-branch consistency loss.}
After the perturbed point cloud $\mathcal{P}'$ is obtained, 
we take the perturbed and original point cloud to the feature extraction networks to extract the point-wise feature $y_i \subset Y  $ and global feature $G$  for basic branch, ${y_i}' \subset Y'  $ and $G'$ for assistant branch.
Then,
we design the dual-branch consistency loss to bring the corresponding feature of different branches closer.
It allows to transmit the discriminative content information from the assistant branch to the basic branch.
Moreover,
to capture multi-scale semantics in the point cloud,
we introduce such consistency losses in different scales,
i.e.,
global consistency loss, 
local consistency loss,
and 
local-to-global consistency loss.

\begin{table*}
	\begin{center}
		\scriptsize
		\setlength\tabcolsep{3.3pt}
		\begin{tabular}{l | l | c| c | cccccccccccccccc }
			\hline
			Method & \%train  &cat. mIoU& ins. mIoU & aero & bag & cap & car & chair & earp. & guitar & {knife} & lamp & laptop & motor & mug & pistol & rocket & { skate.} & table  \\
			\hline \hline
			Kd-Net\cite{Klokov_2017_ICCV} &  & 77.4 & 82.3 & 80.1 & 74.6 & 74.3 & 70.3 & 88.6 & 73.5 & 90.2 & 87.2 & 81.0 & 94.9 & 57.4 & 86.7 & 78.1 & 51.8 & 69.9 & 80.3\\
			PointNet\cite{Charles_2017} &  &80.4 & 83.7 & 83.4 & 78.7 & 82.5 & 74.9 & 89.6 & 73.0 & {91.5} & 85.9 & 80.8 & 95.3 &  65.2 & 93.0 & 81.2 & 57.9 & 72.8 & 80.6 \\
			SO-Net\cite{Li_2018_CVPRso}&  & 80.8 & 84.6 & 81.9 & 83.5 & 84.8 & 78.1 & 90.8 & 72.2 & 90.1 & 83.6 & 82.3 & 95.2 & 69.3 & 94.2 & 80.0 & 51.6 & 72.1 & 82.6\\
			KCNet\cite{Shen_2018_CVPRkc}&  Full & 82.2 &84.7  & 82.8 & 81.5 & 86.4 & 77.6 & 90.3 & 76.8 & 91.0 & 87.0 & {84.5} & 95.5 & 69.2 & 94.4 & 81.6 & 60.1 & 75.2 & 81.3\\
			RS-Net\cite{rs} & (100\%) & 81.4 & 84.9 & 82.7 & 86.4 & 84.1 & 78.2 & 90.4 & 69.3 & 91.4 & 87.0 & 83.5 & 95.4 & 66.0 & 92.6 & 81.8 & 56.1 & 75.8 & 82.2\\
			PointNet++\cite{qi2017pointnet++}& & 81.9 & 85.1 & 82.4 & 79.0 & 87.7 & 77.3 & 90.8 & 71.8 & 91.0 & 85.9 & {83.7} & 95.3 & {71.6} & 94.1 & 81.3 & 58.7 & 76.4 & 82.6  \\
			DGCNN\cite{wang2018dgcnn} & & 82.3 & 85.1 & {84.2} & 83.7 & 84.4 & 77.1 & {90.9} & {78.5} & {91.5} & 87.3 & 82.9 & 96.0 & 67.8 & 93.3 & 82.6 & 59.7 & 75.5 & 82.0 \\
			RSCNN\cite{Liu2019RSCNN} & & 84.0 & 86.2 &  83.5 & 84.8 & 88.8 & 79.6 & {91.2} & 81.1 & 91.6 & 88.4 & 86.0 & 96.0 & 73.7 & 94.1 & 83.4 & {60.5} & 77.7 & 83.6  \\
			\hline
			Ours & Self. (5\%)  & 76.4 & 81.5 & 79.6 & 66.5 & 79.4 & 73.2 & 87.5 & 64.4 & 89.2 & 80.1 & 76.1 & 94.9 & 51.2 & 93.2 & 78.7 & 48.5 & 79.4 & 80.5\\
			Ours & Self. (10\%)  & 76.4 & 81.6 & 79.1 & 73.1 & 76.5 & 70.5 & 87.6 & 63.2 & 89.2 & 82.3 & 77.8 & 95.0 & 52.4 & 90.5 & 78.6 & 48.8 & 75.9 & 80.8\\
			Ours & Self. (50\%)  & 78.8 & 82.5 & 79.0& 78.5& 82.9& 75.0 & 88.0 & 68.8 & 89.8 & 85.4 & 77.1 & 94.9 & 62.6 & 86.8 & 81.6 & 56.7& 72.7 & 82.1\\
			\hline
		\end{tabular}
		
	\end{center}
	\caption{Comparison on ShapeNetPart segmentation task. Average mIoU over
		instances (ins.) and categories (cat.) are reported. 
		Full (100\%): 
		The fully-supervised methods  are trained on the train set of ShapeNetPart (with annotations).
		Self ($R$\%): 
		The self-supervised methods are first pretrained on the train set of ShapeNetPart (\textit{WITHOUT} annotations),
		and then fine-tuned  on only $R$\% train set of ShapeNetPart (the parameters of pre-trained models are fixed).
	}
	\label{Exp_seg_SNPart}
\end{table*}


The global consistency loss $\mathcal{L}_{\text{CG}}$ mainly applies to  preserve the global consistency between the perturbed global representation and basic original representation.
Here we calculate the similarity of  the global features from two branch as:
\begin{equation}
    \mathcal{L}_{\text{CG}} = 1- \cos \left( {G, G'} \right)
\end{equation}  

While the local consistency loss $\mathcal{L}_{\text{CL}}$ acts on point-wise representation, which is used to enhance the feature relevance between the basic branch and assistant  branch.
We can operate the similarity of features of each corresponding point as follows:
\begin{equation}
    \mathcal{L}_{\text{CL}}=-\sum_{i=1}^{N} \log \frac{\exp \left({y}_{i} \cdot {y}_{i}' / \tau\right)}  {\sum_{j=1}^{N}  \exp \left({y}_{i} \cdot {y}_{j}' / \tau\right)}
\end{equation}

Moreover, in order to maximize the lower bound of the mutual information between local  and global representation, then make  the point-wise representation as close as possible to the global representation, we use a local-to-global consistency  loss $\mathcal{L}_{\text{CL2G}}$ to explore the distinct property  by connecting local and global representation  of different branches.
This loss can be formulated as:

\begin{equation}
    \begin{aligned}
        \mathcal{L}_{\text{CL2G}} = & -\sum_{i=1}^{N} \log \frac{\exp \left({y}_{i} \cdot G' / \tau\right)}  {\sum_{k=1}^{B}  \exp \left({y}_{i} \cdot G_k' / \tau\right)} \\ 
        & - \sum_{i=1}^{N} \log \frac{\exp \left({y}_{i}' \cdot G / \tau\right)}  {\sum_{k=1}^{B}  \exp \left({y}_{i}' \cdot G_k / \tau\right)}
    \end{aligned}
\end{equation}
where $B$ is the batch size, $\left\{{G}_{k}, k=1,2, \ldots, B\right\}$ are the global features of different point clouds. 
Finally, the dual-branch consistency loss is $\mathcal{L}_{dual} = \mathcal{L}_{\text{CG}} +\mathcal{L}_{\text{CL}} + \mathcal{L}_{\text{CL2G}} $.

\textbf{Reconstruction Loss.}
Based on our dual-branched framework,
we can extract the global representation $G$ and $G^{'}$ respectively for original and perturbed point clouds.
In order to perform self-reconstruction, a FoldingNet based predictor \cite{yang2018foldingnet} is used to deform the normal 2D grid with $G$ and $G'$ onto the 3D coordinates of the reconstructed point cloud $\hat{\mathcal{P}}$ and $\hat{\mathcal{P}'}$.
We calculate the reconstruction loss of reconstructed point cloud and original point cloud, which is defined as the chamfer distance \cite{fan2017point}:
\begin{equation}
\begin{aligned}
        \mathcal{L}_{\text {recon}}=\sum_{{p} \in \mathcal{P}} \min _{\hat{{p}} \in \hat{\mathcal{P}}}\|\hat{{p}}-{p}\|_{2}+\sum_{\hat{{p}} \in {\hat{\mathcal{P}}}} \min_{{p} \in \mathcal{P}}\|{\hat{{p}}-{p}\|_{2}}   \\
       + \sum_{{p} \in \mathcal{P}} \min _{\hat{{p}}' \in \hat{\mathcal{P}}'}\|\hat{{p}}'-{p}\|_{2}+\sum_{\hat{{p}}' \in {\hat{\mathcal{P}'}}} \min _{{p} \in \mathcal{P}}\|{\hat{{p}}'-{p}\|_{2}}
\end{aligned}
\end{equation}


\textbf{Normal Estimation Loss.}
The point cloud normal vector is a most basic point cloud feature and plays a vital role in many point cloud processing algorithms \cite{Charles_2017,rao2020pointGLR}.
The task of normal estimation requires the establishment of a high level representation  on the surface of the 3D object.
In the process of self-supervised feature learning, we do not need to pursue the accuracy of the estimated normal vectors \cite{Liu2019RSCNN}, but we need to use this task as a self-supervised signal to improve the point-wise level of self-supervised representation. 
We use cosine loss to measure the estimation error:
\begin{equation}
    \mathcal{L}_{\text {normal }}=1-\frac{1}{N} \sum_{i=1}^{N} \cos \left( \hat{{n}}_{i}, {n}_{i}\right)
\end{equation}
where $ \hat{{n}}_{i}$ and $ {{n}}_{i}$ are predicted normal vector and original normal vector.

\begin{table}[]
	\centering
	\begin{tabular}{l | l| c c}
		\hline {\text { Method }}  & Unsupervised  &  {1\%\text{train data}} & {5\%\text{train data }} \\
		\cline { 3- 4 } & pretrain & \text { mIoU (\%) }  & \text { mIoU (\%) } \\
		\hline \hline \text { SO-Net \cite{Li_2018_CVPRso} } & ShapeNet   & 64.0  & 69.0 \\
		\text { PointCapsNet \cite{zhao2019_3DcapsNet}  } & ShapeNet  & 67.0  & 70.0 \\
		\text { MultiTask \cite{hassani2019unsupervised}  }  & ShapeNet & 68.2  & 77.7 \\
		\text { UFF \cite{zhang2020UFF}  } & ShapeNet   & 68.5  & 78.3 \\
		\text { PointContrast \cite{xie2020pointcontrast} } & ScanNet &74.0 &79.9 \\ 
		\text{ Du et. al.} \cite{du2021self} & ShapeNet &76.2 &79.2 \\ 
		\text { Chen et. al. \cite{chen2021shape} } & ShapeNet &  74.1 & 80.1\\
		\hline \textbf { Ours } & ShapeNet  &  \textbf{79.3}  & \textbf{81.2} \\
		\hline
	\end{tabular}
	\caption{Transfer capacity (ShapeNetPart segmentation task).
		Most models are first unsupervisedly pretrained on ShapeNet,
		and then semi-supervisedly finetuned on ShapeNetPart with limited annotations.
		Note that,
		PointContrast is pretrained on larger  ScanNet dataset.
		We can see that,
		our self-supervised model achieves the best performance, 
		especially when the  fine-tuned labeled data is quite limited (e.g., 1\% ShapeNetPart).
		This shows its powerful generalization capacity on limited data. }
	\label{Exp_seg_SN55}
\end{table}

\section{Experiments
\label{Sec_exp}}
This section will introduce the implementation details and the experimental comparisons  for point cloud classification and part segmentation.

\subsection{Implementation Details}
All our models are trained on a single GTX 2080ti GPU with the deep learning library Pytorch \cite{paszke2019pytorch}.
Our model is trained under the Adam \cite{kingma2014adam} optimizer with a basic learning rate of 0.001, and the learning rate is reduced by 0.7 every 20 epoches.
The momentum of the batch normalized \cite{ioffe2015batch} layer starts from 0.9, and then decays at a rate of 0.5 every 20 periods.

As for the self-supervised pre-training of classification, 
we only use three RSConv modules to extract point cloud features in the  extractor network, where the feature does not need to propagate to the original number of points. 
We use the global consistency loss and local-to-global consistency loss to preserve consistency of global fine-grained representations. 
While for the segmentation,
we utilize the  transition-up part \cite{zhao2020pointtransformer} with four layers to obtain more diverse point-wise self-supervised features, where the intermediate  features should be propagated to the original points numbers. 
Focusing on point-wise representation, we can use the local consistency loss  as dual-branch consistency loss. 
As for the evaluation, we can randomly select the training set by category. For the contour-perturbed augmentation  module , we jitter the contour points with normal distributed noise with std of 0.02, which is determined by experimental results in Table \ref{Anysis_std}.

\begin{table}[]
	\centering
	\begin{tabular}{p{55pt}|l|c}
		\hline  & { Method } &    Acc.( \%)  \\
		\hline \hline&  { PointNet \cite{Charles_2017} } & 89.2 \\
		&  { PointNet++ \cite{qi2017pointnet++} }  & 90.5 \\
		& { SO-Net \cite{Li_2018_CVPRso} }  & 92.5\\
		& { PointCNN \cite{li2018pointcnn} }  & 92.5 \\
		Supervised & { DGCNN \cite{wang2018dgcnn} }   & 92.9\\
		& { RSCNN \cite{Liu2019RSCNN} }   & 92.9\\
		& { RSCNN (vote) }   &  93.6\\
		& { DGCNN \cite{wang2018dgcnn} }  & 93.5\\
		& { SO-Net \cite{Li_2018_CVPRso} } & 93.4\\
		& { KPConv \cite{thomas2019kpconv} }  & 92.9\\
		\hline
		& { PointHop \cite{zhang2020pointhop} }  & 89.1 \\
		Unsupervised  & { PointHop++ \cite{zhang2020pointhop} } & 91.1\\        
		(simple)& { PointGLR \cite{rao2020pointGLR} }  & 92.9\\
		& { Chen et. al. \cite{chen2021shape} }& 92.4\\
		& { \textbf{Ours} } & \textbf{92.5}\\
		\hline
		& { FoldingNet \cite{yang2018foldingnet} }   & 88.9\\
		& { PointCapsNet \cite{zhao2019_3DcapsNet} } &  88.9\\
		Unsupervised & { MultiTask \cite{hassani2019unsupervised} } &  89.1\\
		(difficult)& { UFF \cite{zhang2020UFF} } &  90.4\\
		& { \textbf{Ours} } & \textbf{91.9}\\  
		\hline
	\end{tabular}
	\begin{tabular}{l}
	\end{tabular} 
	\caption{Comparison of classification results on ModelNet40 dataset. ``vote" is using the testing voting trick. ``simple" means the self-supervised methods are trained and tested on ModelNet40, while ``difficult" means that the  self-supervised  methods are trained  on  ShapeNet and tested on ModelNet40.}
	\label{Exp_cls_MD40}
\end{table}


\subsection{Point Cloud Part Segmentation}
The purpose of point cloud part segmentation is to predict the part category label of each point in a given point cloud.
We evaluate the features of each point learned by our self-supervised model on the ShapeNetPart dataset \cite{yi2016scalable} which is pre-trained on ShapeNetPart and ShapeNet \cite{chang2015shapenet} dataset.
ShapeNetPart \cite{yi2016scalable} contains 16,881 objects from 16 categories.
 Each object consists of 2 to 6 parts with total of 50 distinct parts among all categories. While ShapeNet \cite{chang2015shapenet} contains 57,000 models across 55 categories. 

As for the training setting, 
in Table \ref{Exp_seg_SNPart}, 
we use the  training set of ShapeNetPart (without annotations) for unsupervised pretraining,
and then use only R\% samples of the training set  for fine-tuning the unsupervised feature.
It can be observed that with only 50\% training annotations, our self-supervised CP-Net can achieve 82.5\% instance mIoU.
Table \ref{Exp_seg_SNPart}  also shows the comparison with fully-supervised models.
The results suggest that our model achieves the mIoU which is only 4.1\% less than the best supervised model.
In addition, to disentangle the performance benefits due to unsupervised training, we trained the fully-supervised  DGCNN \cite{wang2018dgcnn} in Table \ref{Exp_seg_SNPart} with only 5\% training data.
It only achieves 76.6\% instance mIoU, which is worse than our semi-supervised method (81.5\% mIoU with 5\% train set).
This also reflects  that our method has powerful generalization capacity on limited data.

In Table \ref{Exp_seg_SN55},
we use the whole train set of ShapeNet (without annotations) for pretraining and use 1\%/5\% train set of ShapeNetPart for  fine-tuning.
These two settings follow \cite{hassani2019unsupervised, Li_2018_CVPRso,zhang2020UFF,zhao2019_3DcapsNet}, to make fair comparison.
In order to assess the transferability of the self-supervised methods, we train our model on ShapeNet \cite{chang2015shapenet} and then evaluate it on ShapeNetPart.
The results are shown in Table \ref{Exp_seg_SN55}, our method has a better performance than other methods on the transfer learning to ShapeNetPart.

Moreover, 
in the setting of GraphTER \cite{gao2020graphter},
they use 100\% train set of ShapeNetPart (with labels) for supervised fine-tuning (same pretraining like us). 
In this case,
our setting is actually more challenging with less data in fine-tuning.
Under the same setting of GraphTER \cite{gao2020graphter},
our method (83.2\%) outperforms GraphTER (81.9\%) on ShapeNetPart segmentation task.

\begin{table}[]
	\centering
	\begin{tabular}{l | l | c }
		\hline  & { Method } &   { Acc.(\%)} \\
		\hline \hline&  { 3DmFV \cite{ben20183dmfv} }  & 73.8 \\
		&  { PointNet \cite{Charles_2017} }  & 79.2 \\
		Supervised &  { SpiderCNN \cite{xu2018spidercnn} }  & 79.5 \\
		& { DGCNN \cite{wang2018dgcnn} }  & 86.2\\
		& { PointCNN \cite{li2018pointcnn} }  & 85.5 \\
		& { GDANet \cite{xu2020GDA} } & 88.1\\
		& { Point-BERT \cite{yu2021point} } & 88.1\\
		\hline
		Unsupervised & { PointGLR \cite{rao2020pointGLR}}  &86.9\\
		& {\textbf{ Ours}  } & \textbf{87.9}\\
		\hline
	\end{tabular}
	\caption{Comparison of classification results on real-world  ScanObjectNN dataset (OBJ ONLY). }
	\label{Exp_cls_ScanOBJ}
\end{table}

\begin{figure*}[t]
	\begin{center}
		\includegraphics[width=1\linewidth]{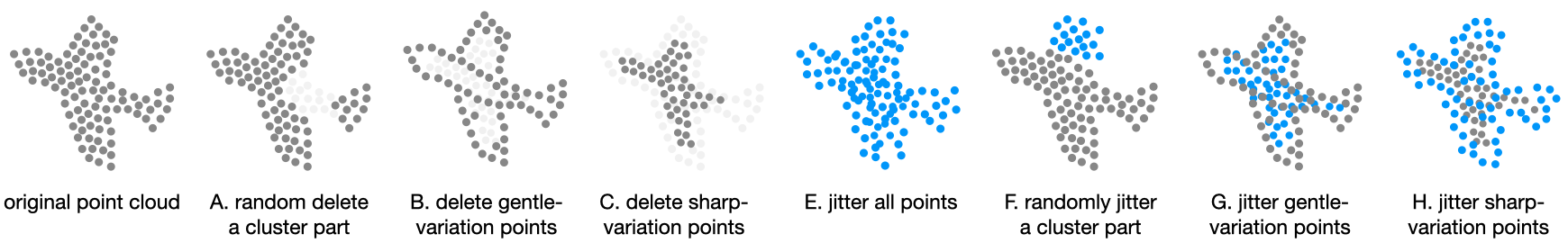}
	\end{center}
	\caption{Different perturbation manners for the assistant branch (corresponding to Table \ref{Anysis_contrast_pointcloud}).  
	}
	\label{vis_perturb}
\end{figure*}

\subsection{Unsupervised Point Cloud Classification}
For the unsupervised classification, 
we first obtain the self-supervised shape features from the ModelNet40 \cite{wu20153d} and ScanObjectNN \cite{uy2019revisiting} dataset using self-supervised pre-trained model.
Then we use a linear SVM \cite{cortes1995support} to   classify self-supervised shape features.
ModelNet40  \cite{wu20153d} is a benchmark dataset for shape classification.
It contains 9,843 training samples, 2,468 testing samples and 40 object categories, where the points are sampled from CAD models.
ScanObjectNN  \cite{uy2019revisiting} is a real-world dataset, where 2,902 3D objects are extracted from scans.
In our classification experiments, we sample 1,024 points for each point cloud for training and evaluation.
All our results are measured using a single view without using the multi-view voting trick to show the neat performance of different models.
Surface normal vectors are used to provide self-supervised signals for our models trained on ModelNet40 and we did not use it as input.
For the models trained on ScanObjectNN we do not use the normal vectors, because normal vector  of real-world data is not accurate enough.


\begin{table}[]
	\centering
	\begin{tabular}{l|c}
		\hline
		Our CP-Net & mIoU(\%) \\\hline \hline
		only basic branch & 78.4\\
		only assistant branch & 78.2\\
		basic branch and assistant branch & 81.5\\
		\hline  
	\end{tabular}
	\caption{Ablation study of the branches. We report the mIoU on ShapeNetPart semi-supervised segmentation with 5\% train data, where the self-supervised features are learned from ShapeNetPart.}
	\label{Anysis_branch}
\end{table}

\begin{table}[t]
	\centering
	\begin{tabular}{c| l|c}
		\hline 
		& Perturbation Manners & mIoU \\ \hline \hline
		A & Randomly delete a cluster part & 78.7  \\
		B & Delete content points & 80.8 \\
		C & Delete contour points & 81.0 \\
		D & Randomly delete contour or content points  & 79.9  \\
		E & Jitter all points & 79.6 \\
		F & Randomly jitter a cluster part  & 78.8 \\
		G & Jitter content points & 80.8 \\
		H & Jitter contour points & 81.5 \\
		I & Randomly jitter  contour or content points& 81.1 \\
		\hline
	\end{tabular}
	\caption{Ablation Study of the perturbation of contour-perturbed augmentation  module. We report the mIoU on ShapeNetPart semi-supervised segmentation with 5\% train data, where the self-supervised features are learned from ShapeNetPart.}
	\label{Anysis_contrast_pointcloud}
\end{table}

\textbf{Unsupervised learning on ModelNet40.}
As shown in Table \ref{Exp_cls_MD40}, we compare the performance of unsupervised classification methods and fully-supervised classification methods.
The top part (Supervised) is the result of the fully-supervised SOTA methods, the middle part (Unsupervised simple) is the classification result of self-supervised feature learning on ModelNet40, and the bottom part (Unsupervised difficult) is the unsupervised transfer learning from the ShapeNet to the ModelNet40 dataset.

As for learning on ModelNet40, many studies \cite{wang2018dgcnn, li2018pointcnn,Liu2019RSCNN} have shown that the performance of ModelNet40 has been gradually saturated. 
Our method achieves a comparable performance 92.5\% with the SOTA unsupervised method PointGLR\cite{rao2020pointGLR}, and more notably, our method can even achieve very competitive results compared to SOTA supervised methods (92.9\% without voting trick) in an unsupervised manner.
This evidence indicates that our method can discover global semantic representation shared in different kinds of point clouds. 

For better evaluation and further explore the generalization ability of the learned representation,
we use a more challenging transfer setting (Table \ref{Exp_cls_MD40}: bottom), 
we test our method with transfer learning from ShapeNet to ModelNet40. 
i.e., ShapeNet (unsupervised pretraining) + ModelNet40 (SVM for evaluating the unsupervised representation).
Our method largely outperforms the SOTA approaches.
It shows that the unsupervised features from our method are more generic than other methods.

\begin{table}[t]
	\centering
	\begin{tabular}{c|c c c c}\hline
		\# Jittered Points & 512 &1024 &1536  & 2048\\
		\hline \hline
		mIoU (\%)& 80.9 &81.5 &81.3 & 79.6\\
		\hline
		
	\end{tabular}
	\caption{Evaluation of the mIoU when increasing the number of jittered points of the contour points (all points number is 2048). We report the mIoU on ShapeNetPart  segmentation with 5\% train data.}
	\label{Anysis_points}
\end{table}

\begin{table}[t]
	\centering
	\begin{tabular}{c|c c c}\hline
		STD & 0.01 &0.02 &0.03 \\
		\hline \hline
		mIoU (\%)& 80.5 &81.5 &80.7\\
		\hline
		
	\end{tabular}
	\caption{Experiments of the std of the normal distributed noise. We report the mIoU on ShapeNetPart  segmentation with 5\% train data.}
	\label{Anysis_std}
\end{table}

\textbf{Unsupervised Learning on ScanObjectNN.}
In order to verify the effectiveness of our method more comprehensively, we conduct the same unsupervised classification task on the ScanObjectNN \cite{uy2019revisiting}.
ScanObjectNN is used to investigate the robustness to noisy objects with deformed geometric shape and non-uniform surface density in the real world.
We adopt our model on the $OBJ\_ONLY$ (simplest variant of the dataset).
The results are summarized in Table \ref{Exp_cls_ScanOBJ},
our method achieves the comparable accuracy with the fully-supervised methods, and this proves that our method has strong practicality in the real world data.

\section{Network Analysis
\label{Sec_analysis}}
In this section, we first introduce the ablation studies of our framework. 
Second, we analyze the details of  contour perturbation module.
Then we further present the analysis of the normal estimation loss, self-reconstruction loss and the dual-branch consistency loss.
Moreover, we also  analyze the robustness of our method on sampling density.

\subsection{Ablation Studies of the Network Architecture}
In order to examine the effectiveness of our designs, we conduct  architecture ablation studies  based on our framework. 
In Table \ref{Anysis_branch}, we  conduct the comparison of the branches in our framework, 
where we only use the basic branch or assistant branch, there is a large gap on the performance compared with the fully complete dual branches.
It can be concluded that the feature distinction of  contour components (basic branch) and semantic content components (assistant branch) are both important in the self-supervised feature representation.
More importantly, with the consistency learning between the two branches, we can improve the performance  largely by enhancing the representation relevance between the semantic content information and structural contour information.




\subsection{Analysis of Contour-perturbed Augmentation  Module
	\label{sec_analysis_SPM}
}
To explore an effective way to perturb the point cloud for self-supervised feature learning, as Table \ref{Anysis_contrast_pointcloud}  and Figure \ref{vis_perturb} shows, we  design a variety of ways to perturb the point cloud to the assistant branch. We first consider the aggregation part information of the original point cloud. Specially, we use the non-negative matrix factorization method \cite{lee1999learning} to extract similar clustering effects from the original point cloud, and then randomly select one cluster as the cluster part mentioned in Model A and F. 
However, no matter jitter or delete a cluster part randomly, the results are not particularly ideal, because there extends a large gap between the parts obtained by clustering and the ground truth, and it is might mislead the network to learn some distracting information. To verify the effectiveness of contour-perturbed augmentation  module, we conduct a series of comparative experiments (Model B,C,D,G,H,I). Model C indicates that the contour points 
have certain useful information; Model B and D show that delete the content  points can harm the performance ($0.7\%\downarrow$). 
Model G, H and I verify that jittering contour points can highlight the learning of holistic and generic information.

Based on Model H, we further analyze the perturbation details. Table \ref{Anysis_points} shows the  mIoU evaluation when increasing the number of jittered points of the contour points, when we use 1024 points ($N/2$), we get the best performance. 
As for the normal distributed noise, we conduct a series of experiments based on the std of the normal distributed noise, which is shown in Table \ref{Anysis_std}.


\begin{table}[t]
	\centering
	\begin{tabular}{p{25pt} c c c | c }
		\hline Model & $\mathcal{L}_{normal} $  &  $\mathcal{L}_{recon} $ &$\mathcal{L}_{dual}$ & {mIoU(\%)} \\
		\hline \hline 
		(a) & \checkmark & & & 71.7 \\
		(b) &  & \checkmark & & 75.9\\
		(c) &  &  & \checkmark & 79.8\\
		(d) & \checkmark & \checkmark & & 78.4\\
		(e) & \checkmark &  & \checkmark & 81.2\\
		(f) &  & \checkmark & \checkmark & 80.8\\
		(g) & \checkmark & \checkmark & \checkmark &81.5  \\
		\hline  
	\end{tabular}
	\caption{Ablation study of losses. We report the mIoU on ShapeNetPart semi-supervised segmentation with 5\% train data, where the self-supervised features are learned from ShapeNetPart.}
	\label{Anysis_Ablation}
\end{table}

\begin{table}[t]
	\centering
	\begin{tabular}{p{25pt} c c c | c }
		\hline Model & $\mathcal{L}_{CL} $  &  $\mathcal{L}_{CG} $ &$\mathcal{L}_{CL2G}$ & {mIoU(\%)} \\
		\hline \hline 
		a & \checkmark & & & 81.5 \\
		b & \checkmark & \checkmark & & 80.5\\
		c & \checkmark & \checkmark & \checkmark & 80.2\\
		\hline  
	\end{tabular}
	\caption{Ablation study of dual-branch consistency loss for segmentation downstream task. We report the mIoU on ShapeNetPart semi-supervised segmentation with 5\% train data, where the self-supervised features are learned from ShapeNetPart.}
	\label{Anysis_Ablation_supp}
\end{table}

\begin{table}[t!]
	\centering
	\begin{tabular}{p{25pt} c c c | c }
		\hline Model & $\mathcal{L}_{CG} $  &  $\mathcal{L}_{CL} $ &$\mathcal{L}_{CL2G}$ & {Acc.(\%)} \\
		\hline \hline 
		A & \checkmark & & & 91.3 \\
		B & \checkmark & \checkmark & & 91.5\\
		C & \checkmark & \checkmark & \checkmark & 92.5\\
		\hline  
	\end{tabular}
	\begin{tabular}{l}
		\small
	\end{tabular} 
	\caption{Ablation study of consistent loss for clsssification downstream task. We report the accuracy on ModelNet40 test set.}
	\label{Anysis_Ablation_supp_cls}
\end{table}

\subsection{Analysis of Network Losses}

\begin{figure}[t]
	\begin{center}
		\includegraphics[width=0.66\linewidth]{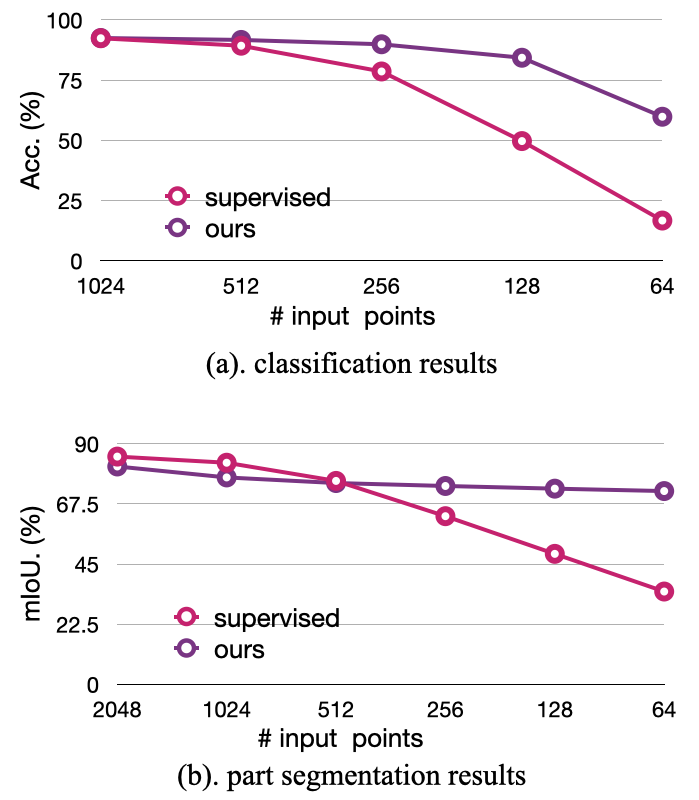}
	\end{center}
	\caption{Sampling density robustness test compared to the supervised version on classification and part segmentation. 
		(a). Test results on ModelNet40 of using sparser points as the input to a model trained with 1,024 points. (b) Test results on ShapeNetPart of using sparser points as the input to a model trained with 2,048 points. 
	}
	\label{Anysis_robustness}
\end{figure}



In Table \ref{Anysis_Ablation}, we report the mIoU of  segmentation with 5\% train data, and the self-supervised representation is learned from ShapeNetPart dataset.
Model (a), (b) and (d) are based on our baseline model which is training with the basic branch, without assistant branch.
Model (a) can be viewed as a variant of FoldingNet \cite{yang2018foldingnet}, which is trained by self-reconstruction loss only and get a low segmentation mIoU of 71.7\%, while model (b) and (d) make a slight improvement with normal estimation loss, because normal estimation is a point-wise task, this self-supervised signal can affect the self-supervised feature for each point, thereby improving the performance of part segmentation. 
When we only use dual-branch consistency loss without reconstruction and normal estimation loss, we can still get a result of 79.8\%. 
Compared with model (a), (b) and (d), model (c) shows that the contour-perturbed augmentation  module with dual-branch consistency loss has a great improvement in performance.  
Based on dual-branch consistency loss, we add the normal estimation module, self reconstruction module to model (c), which can be indicated as  model (e), (f) and (g).
They boost significant improvements, and the best performance is 81.5\% of model (g).

Here, we further analyze the dual-branch consistency losses. 
These losses  can be used to minimize the feature distance between the point clouds from basic branch and assistant branch, leading the self-supervisor to obtain diverse and effective representation. 
Due to different attentions on point cloud representation of the different downstream tasks, here we conduct some ablation studies of the dual-branch consistency losses for classification downstream task and part segmentation downstream task.
 As Table \ref{Anysis_Ablation_supp_cls} shows, for classification, global consistency loss and local to global consistency loss play a critical role in performance improvement. 
 For part segmentation, which focus more on point-wise local representation, the local consistency loss is more critical as Table \ref{Anysis_Ablation_supp} indicates. Thus, for  pre-training of part segmentation task, we can use the local consistency loss as  dual-branch consistency loss.

\begin{figure}[t]
	\begin{center}
		\includegraphics[width=0.9\linewidth]{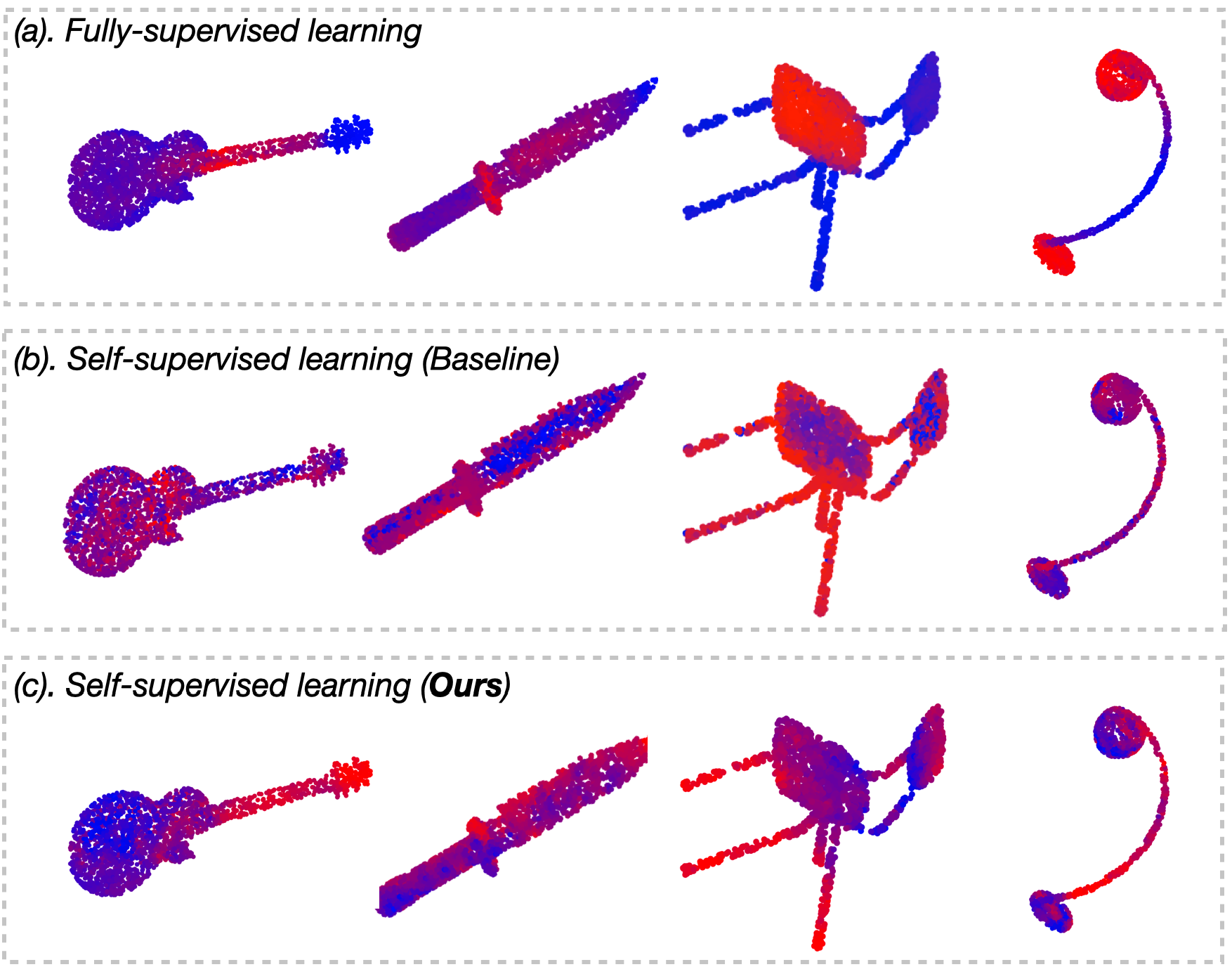}
	\end{center}
	\caption{ Comparison of Feature representation. We show the supervised version (a),  self-supervised baseline (b) and our self-supervised CP-Net (c). The point color refers to the activation value of each point which is obtained by averaging all the entries in the feature vector.
	}
	\label{Anysis_vis_feats}
\end{figure}

\subsection{Robustness Analysis}
Figure \ref{Anysis_robustness} shows the robustness of our method on sampling density compared to the supervised version.
Following \cite{rao2020pointGLR}, we use sparse points (1024, 512, 256, 128, 64) as the input of classification model and part segmentation model for testing. 
For classification (Figure \ref{Anysis_robustness} (a)), we feed sparse point clouds  to the  model trained with 1024 points, and obtain the self-supervised feature, then use a linear SVM to perform the classification results.
We can see that our self-supervised classification  model is much more robust than the supervised model on sampling density. 
Even we use 128 points for testing, the accuracy can achieve to 84.3\%.
For part segmentation  (Figure \ref{Anysis_robustness} (b)), we also feed the point clouds with different densities to a segmentation model which is trained with 2048 points.
Then we use the same setting with Table \ref{Exp_seg_SNPart} to perform the segmentation results. 
It can be concluded that our self-supervised method is more robust to the  point cloud density.

\section{Visualization}

\subsection{Reasonable Segmentation }

\begin{figure}[t]
	\begin{center}
		\includegraphics[width=0.9\linewidth]{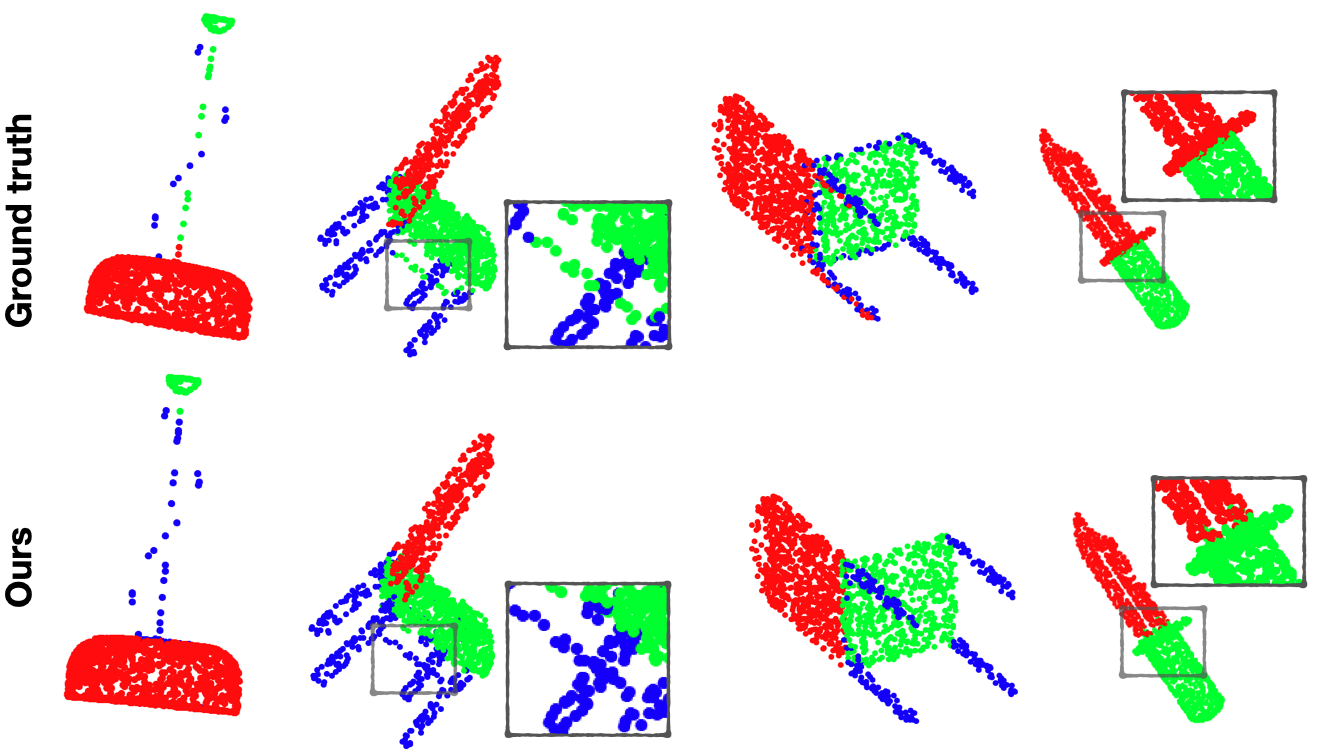}
	\end{center}
	\caption{Point cloud reasonable segmentation results. The first row shows the ground truth part labels. The second row shows the predict labels by our method. Even though the ground truth are noisy, our method can get reasonable segmenting results of the objects.
	}
	\label{fig_supp_1}
\end{figure}

In Table \ref{Exp_seg_SNPart} and \ref{Exp_seg_SN55}, we show the part segmentation results of our SPR-Net on ShapeNetPart dataset \cite{yi2016scalable}. 
The mIoU is an important statistical evaluation metric which indicates statistical overall performance on test dataset.
Occasionally, the ground truth  may be confused in some situations.
In order to show our results more comprehensively, here we visualize some test segmentation results of our method with the ground truth in Figure \ref{fig_supp_1}. 
Although our results are different from the manually annotation, they are both reasonable manners of segmenting the objects. Our method divides the lamp bracket and lamp rope into the same category, 
while the ground truth labels consider that bases and brackets  belong to the same category,
The situation is similar for shank, chair leg brackets ...

\subsection{Feature Representation Visualization}
In order to have an intuitive understanding of our models, we colored the point cloud according to the point cloud feature response on the test set of ShapeNetPart in Figure \ref{Anysis_vis_feats}. 
The points (which belong to the same part) have similar activation. 
It can be observed the fully supervised feature representations can be well distinguished  among the different parts.
The common self-supervised feature representation from baseline model reflect confusion among the different parts.
Compared with the baseline, our self-supervised CP-Net shows more distinction which verifies that our model can  learn semantic content more effectively. 
Our method still needs to be improved in some object details, such as the lamp.

\section{Conclusion}
We propose a dual-branched CP-Net for point cloud self-supervised learning. 
Equipped with contour-perturbed augmentation  module  and dual-branch consistency loss, 
our CP-Net  not only preserves the  discriminative representation on easy-to-learned structural contour information, 
but also  extract the semantic content information which
is hard to learn by self-supervisors.
Extensive experiments have shown the performances, transferability and decent robustness of our CP-Net.


{
\bibliographystyle{IEEEtran}
\bibliography{bibfile}
}


\ifCLASSOPTIONcaptionsoff
  \newpage
\fi

\end{document}